\documentclass[]{IEEEtran}
\IEEEoverridecommandlockouts

\usepackage{cite}
\usepackage{amsmath,amssymb,amsfonts,dsfont}
\usepackage{algorithm}
\usepackage{algpseudocode}
\usepackage{varwidth} 
\usepackage{graphicx}
\usepackage{textcomp}
\usepackage{xcolor}
\usepackage{siunitx}
\usepackage{pgfplots}
\pgfplotsset{compat=1.18}
\usepackage{booktabs}
\usepackage[hidelinks]{hyperref}
\usepackage{bm} 
\usepackage{acronym} 
\usepackage[bb=boondox]{mathalfa} 
\def\BibTeX{{\rm B\kern-.05em{\sc i\kern-.025em b}\kern-.08em
    T\kern-.1667em\lower.7ex\hbox{E}\kern-.125emX}}

\newacro{GP}{Gaussian process}
\acrodefplural{GP}[GPs]{Gaussian processes}
\newacro{TLBO}{transfer learning Bayesian optimization}
\newacro{ML}{Machine learning}
\newacro{MTGP}{multi-task Gaussian processes}
\newacro{BO}{Bayesian optimization}
\newacro{RGPE}{ranking-weighted Gaussian process ensemble}
\newacro{SOTA}{state of the art}
\newacro{SE}{squared exponential}

\DeclareMathOperator*{\argmin}{arg\,min}

\def\w{\bm w}
\def\x{\bm x}

\begin{document}

\title{Sample-Efficient Bayesian Transfer Learning for Online Machine Parameter Optimization} 

\author{\IEEEauthorblockN{Philipp Wagner$^1$, Tobias Nagel$^1$, Philipp Leube$^2$ and Marco F. Huber$^{1,3}$} \\
\IEEEauthorblockA{$^1$\emph{Fraunhofer Institute for Manufacturing Engineering and Automation IPA}, Stuttgart, Germany \\
$^2$\emph{TRUMPF SE + Co. KG}, Ditzingen, Germany \\
$^3$\emph{Institute of Industrial Manufacturing and Management IFF}, University of Stuttgart, Germany\\
\{philipp.wagner $|$ tobias.nagel\}@ipa.fraunhofer.de, philipp.leube@trumpf.com, marco.huber@ieee.org}
}

\maketitle

\begin{abstract}
Correctly setting the parameters of a production machine is essential to improve product quality, increase efficiency, and reduce production costs while also supporting sustainability goals. Identifying optimal parameters involves an iterative process of producing an object and evaluating its quality. Minimizing the number of iterations is, therefore, desirable to reduce the costs associated with unsuccessful attempts. This work introduces a method to optimize the machine parameters in the system itself using a \ac{BO} algorithm. By leveraging existing machine data, we use a transfer learning approach in order to identify an optimum with minimal iterations, resulting in a cost-effective transfer learning algorithm. We validate our approach on a laser machine for cutting sheet metal in the real world.
\end{abstract}


\section{Introduction}
\let\thefootnote\relax\footnotetext{This paper is accepted in IEEE Conference on Artificial Intelligence,
2025. IEEE copyright notice ©2025 IEEE. Personal use of this material is
permitted. Permission from IEEE must be obtained for all other uses, in any
current or future media, including reprinting/republishing this material for
advertising or promotional purposes, creating new collective works, for resale
or redistribution to servers or lists, or reuse of any copyrighted component
of this work in other works.} Optimizing parameters in manufacturing processes is critical for improving product quality, increasing productivity, and reducing costs. These advancements are essential for addressing both economic and sustainability goals within modern production systems~\cite{chia2022process}. In a production environment, parameter optimization is typically performed by running an optimization procedure within a simulation to explore various scenarios~\cite{LI2020117314}. 
\ac{ML} plays an increasingly pivotal role in parameter optimization by allowing more accurate predictions and real-time adjustments~\cite{weichert2019review}. \ac{ML} methods can analyze large datasets to identify optimal settings, uncover hidden patterns, and automate decision-making processes. However, \ac{ML} algorithms often require large amounts of data to perform effectively, limiting their applicability in scenarios involving new products or slightly altered production conditions. In such cases, the lack of sufficient data can result in suboptimal or unreliable model performance~\cite{dou2023machine, kraljevski2023machinelearningsmalldata, xu2023small}. At the same time, classical model-based optimization is often either impossible due to the process complexity or uneconomic if the lot size is too small. Transfer learning and \ac{BO} offer potential solutions in such cases, as they address the challenges posed by limited data and altered manufacturing scenarios. Transfer learning allows models to leverage knowledge from related tasks~\cite{zhuang2021tlsurvey}, while \ac{BO} aims to efficiently explore parameter spaces with minimal data~\cite{garnett_bayesoptbook_2023, sano2020application}, making these approaches well suited for optimizing new or modified production processes.
However, many approaches in the area of transfer learning are hardly applicable, as they require huge amounts of data in the source domains~\cite{bommasani2021opportunities, subramanian2024towards, zhang2023large}.

\begin{figure}[tbp]
\centerline{\includegraphics[width=0.5\linewidth, angle=270]{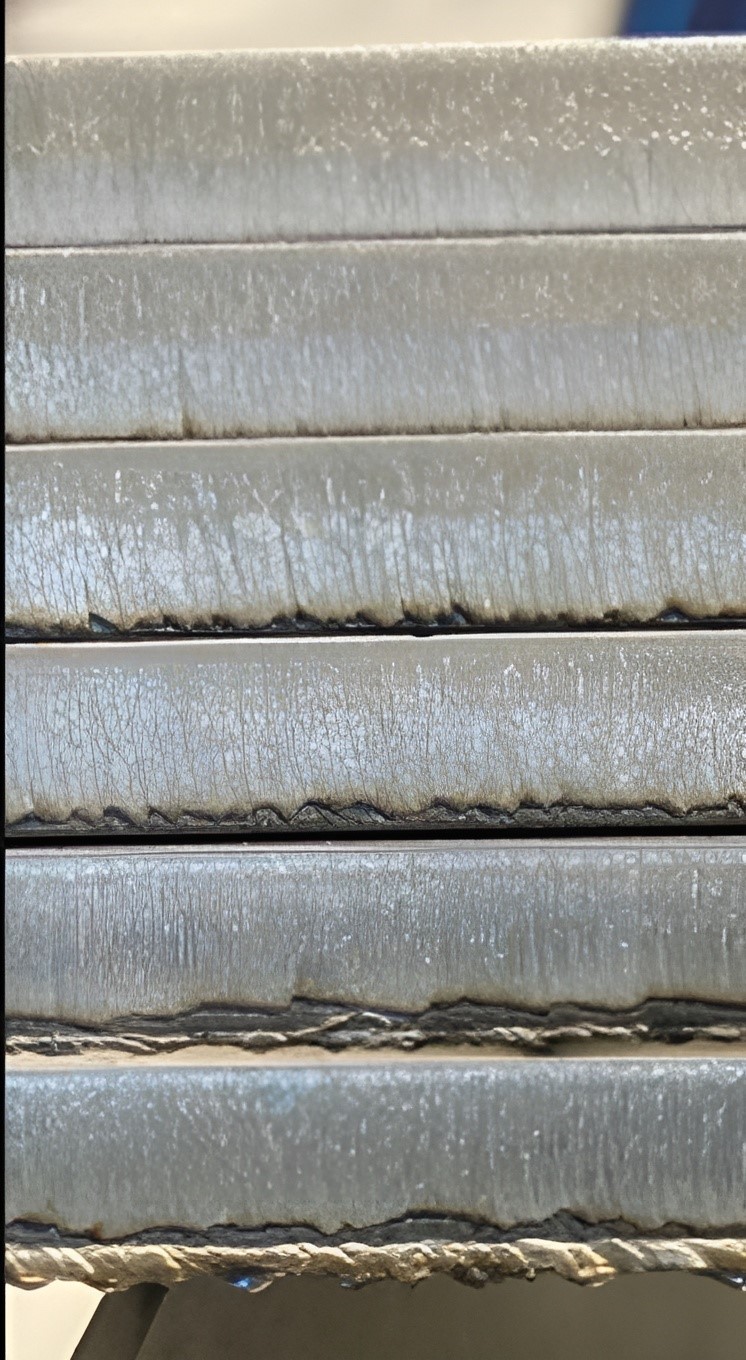}}
\caption{Demonstration of our machine parameter optimization in thermal cutting with six iterations. 
Iteratively adjusting the machine parameters reduces the burr from initially high (left) to barely visible (right).}
\label{fig:schnittkanten}
\end{figure}

The main issue in optimizing a production machine is the infeasibility of conducting numerous optimization iterations, as each iteration in the process produces waste and consumes energy, material, and time. Here, \emph{iteration} describes the steps of adjusting machine parameters, executing one production cycle, and measuring the quality of the produced item. This paper addresses this critical problem by proposing a method that minimizes the number of iterations required for machine parameter optimization by re-using available data from earlier similar production procedures. Our approach is based on the concept introduced in \cite{feurer2018scalable}, where we adapt it to the specific demands of production environments. The proposed method leverages knowledge from similar datasets, where a limited amount of data is available, to efficiently guide the optimization process. By incorporating this prior knowledge, we seek to improve performance under new conditions with minimal experimental effort. To achieve this, we construct an ensemble model that combines information from previous production processes and adapts it to the current optimization process. The ensemble assigns weights to its components based on their relevance to the process under consideration, prioritizing data from similar processes. This weighting mechanism requires only a small amount of additional data from the current process, making it highly efficient. By effectively reusing existing data, our method significantly reduces the need for costly and time-consuming iterations while maintaining high optimization performance. 

To prove the effectiveness of our method, we validate the algorithm on a manufacturing use case in thermal laser cutting. In this use case, a laser cutter of varying power aims to cut sheet metal with high-quality cutting edges both on a surrogate model of the cutting process and on the actual machine hardware. We aim to perform the optimization in as few steps as possible to obtain metal sheets with as few burr as possible as seen in Fig.~\ref{fig:schnittkanten}.

The structure of this work is as follows: we begin with a brief overview of \ac{BO}, transfer learning, and the fundamentals of laser cutting in Sec.~\ref{sec:relatedwork}. In Sec.~\ref{sec:problem}, we provide a detailed problem formulation and introduce our approach, which is based on \ac{RGPE}~\cite{feurer2018scalable}. Then, in Sec.~\ref{sec:TLBO}, we describe the process of collecting data across various scenarios and the training of models using the available data. In Sec.~\ref{sec:evaluation}, we conduct experiments to evaluate the performance of our method. The paper concludes with final remarks and an outlook for future work.

\section{Related Work}
\label{sec:relatedwork}

In this section, we focus on related work in the field of \ac{TLBO} as well as prior work regarding metal sheet optimization in thermal cutting.

The method used in this work is based on the idea of \ac{RGPE}~\cite{feurer2018scalable}. 
In their work, the authors propose a ranking loss to determine ensemble weights for trained \acp{GP} of known tasks to improve performance on the (unknown) target task. While this approach works well in general, it struggles with use cases where tasks are similar but are located in different areas both in the parameter domain and in the optimization target domain. Another approach is to learn a covariance function over tasks in addition to an input-dependent covariance function in \ac{GP} regression~\cite{Bonilla2007MTGP}. This approach is called \ac{MTGP} and can be used in \ac{BO} to leverage data from similar tasks~\cite{swersky2013_MTGPBO}. In~\cite{tighineanu2022transfer}, the authors propose a method based on  hierarchical \acp{GP}~\cite{golovin2017google}. Overviews about \ac{TLBO} in general are given in~\cite{Xilo2023survey} and~\cite{bai2023tlbosurvey}. Learning a covariance function over tasks can improve the optimization performance. However, these approaches are computationally less efficient than the aforementioned ensemble methods.

In~\cite{Tatzel2021_1000137690} and~\cite{Menold2024laserBO}, the authors propose and discuss several approaches for the use of \ac{ML} in thermal cutting, especially in the field of quality prediction and optimization. In~\cite{Menold2024laserBO}, the authors use \ac{BO} from scratch to optimize the cutting parameters for laser cutting, welding, and polishing. The authors in~\cite{michalowski2023advanced} use \ac{BO} in other metal processing use cases such as laser welding and milling.
A bottleneck for an automated parameter optimization in thermal cutting is the objective evaluation of cutting edges. There exist several different works on predicting different quality metrics of thermal cutting of metal edges~\cite{stahl2023comprehensive, Stahl2024drag, stahl2019quick}. This allows for the use of basic RGB cameras during quality assessment. The automated and accurate quality assessment of thermal cutting edges is a foundation of our work.

\section{Problem Formulation}
\label{sec:problem}
In the following, we give an exact problem definition and define the optimization setup we are dealing with. We assume that prior to the optimization, there exist datasets of $M$ tasks $\mathcal{D}_1, \dots, \mathcal{D}_M$ with $\mathcal{D}_m = \{(\bm x_i^m, y_i^m)\, |\, i=1,2,\ldots,N_m\}$, $m=1,2,\ldots, M$. Here, $\bm x_i^m$ describes the $i$-th machine parameter vector of the $m$-th task, $y_i^m$ describes the respective product quality, and $N_m$ denotes the number of the respective data pairs. We call $\mathcal{D}_1, \dots, \mathcal{D}_M$ the data of the \emph{source tasks}. The so-called \emph{target task} is the process to be optimized, where no prior data is available; therefore, the respective data set $\mathcal{D}_t = \{(\bm x_i^t, y_i^t)\}_{i=1}^{N_t}$ is empty at the beginning. The superscript $t$ denotes data from the target process, which is due to optimization. 

In \ac{BO} the goal is to solve the general optimization problem
\begin{equation}
    \bm x^\ast = \argmin_{\bm x \in \mathds{R}^n} f^t(\bm x) \label{eq:bo}
\end{equation}
with the minimizer $\bm x^\ast \in \mathds{R}^n$ of length $n \in \mathbb{N}$ and an unknown function $f^t$ that describes the relationship between input and output data of $\mathcal{D}_t$ according to
\begin{equation} \label{eq:03_gt_f}
    y_i^t = f^t(\bm x_i^t) + \epsilon_i
\end{equation}
 with Gaussian noise $\epsilon_i \sim \mathcal{N}(0, \sigma_i)$. $N_t$ is the number of samples. We utilize the available knowledge, encoded in the datasets $\mathcal{D}_1,\dots, \mathcal{D}_M$, to simplify the optimization problem and achieve a faster convergence to the minimum. In thermal cutting $\bm x_i$ represents a set of process parameters, $y_i$ is a measure of edge quality, and $f_t$ is the unknown relationship between the parameters and the quality. Furthermore, in our problem setup we assume that the input space is constrained by $\bm x \in [\bm x_\mathrm{min}, \bm x_\mathrm{max}]$. This assumption is very common in many real-world applications where the parameter space is limited either by physical properties of the manufacturing process or by other outer constraints. In laser cutting, this assumption arises from the fact that cut interruptions occur if a certain process parameter window is left \cite{leiner2023cut}. However, similar boundary conditions arise in many industrial use cases where the process window is exited. To include other soft constraints, Eq.~\eqref{eq:bo} can be modified by adding a term $\lambda \cdot g(\bm x)$, where $g(\bm x)$ encodes the constraint and $\lambda$ is a non-negative weighting factor.

\section{Transfer Learning for Bayesian Optimization}
\label{sec:TLBO}
In the following, we first introduce the concept of \iac{GP}. Afterwards, we introduce our method to perform a data-efficient \ac{BO} by using knowledge of related processes.

\subsection{Gaussian Process}
\Iac{GP} is a stochastic regression method, which uses a prior distribution over functions that are transformed into a posterior distribution after obtaining some measurement data~\cite{murphy2012machine}. The target is to identify an unknown function $f$, similar to Eq.~\eqref{eq:03_gt_f}. \Iac{GP} is defined by a mean function $m(\bm x) = \mathbb{E}(f(\bm x))$ and a covariance or kernel function $k(\bm x,\bm x') = \mathbb{E}((f(\bm x) - m(\bm x)) \cdot  (f(\bm x') - m(\bm x'))$, which gives the similarity between two points $\bm x$ and $\bm x'$. A frequently chosen kernel is the \ac{SE} kernel, which is given by $ k(\bm x,\bm x') = \sigma^2 \exp \left(-\frac{1}{2l^2} (\bm x - \bm x')^2 \right)$ with two hyper-parameters $\sigma^2$ and $l$ that specify the vertical variation and the length scale, respectively. After obtaining multiple observations $\bm X = [\bm x_1,\ldots, \bm x_N]$, the \ac{GP} defines a joint Gaussian according to
\begin{equation}
    p(\bm f | \bm X) = \mathcal{N}(\bm f | \bm \mu, \bm K)
\end{equation}
with kernel matrix $\bm K$ with elements $K_{ij} = k(\bm x_i, \bm x_j)$ and $\bm \mu = (m(\bm x_1), \ldots, m(\bm x_N))$. By defining test data points $\bm X_*$, the \ac{GP} can predict the posterior of the function according to
\begin{equation}
    \begin{pmatrix}
        \bm f \\
        \bm f_*
    \end{pmatrix} \sim \mathcal{N}\left(
        \begin{pmatrix}
            \bm \mu \\ \bm \mu_*
        \end{pmatrix} ,
        \begin{pmatrix}
        \bm K & \bm K_* \\
        \bm K_*^\mathrm{T} & \bm K_{**}
        \end{pmatrix}
        \right)~,
\end{equation}
where $\bm \mu_*$ denotes the mean vector of the test data, $\bm K_*$ denotes the covariance matrix between the observations and the test data, and $\bm K_{**}$ denotes the covariance matrix of the test data with itself. The mean value and covariance function of the posterior are then calculated according to
\begin{equation}
    p(\bm f_* | \bm X_*, \bm X, \bm f) = \mathcal{N}(\bm f_* | \bm \mu_*, \bm \Sigma_*)
\end{equation}
with $\bm \mu_* = \bm \mu(\bm X_*) + \bm K_*^\mathrm{T} \bm K^{-1} (\bm f - \bm \mu(\bm X))$ and $\bm \Sigma_* = \bm K_{**} - \bm K_*^\mathrm{T} \bm K^{-1} \bm K_*$.
\Iac{GP} is a non-parametric method, which means that in general there are no parameters to be trained in order to obtain a regression model, but the \ac{GP} is defined solely by the data. It is very common, however, to adjust the kernel hyper-parameters in order to obtain a better fit.

    \subsection{General Description}
    The general description of our method is as follows: We use the available data sets $\mathcal{D}_1,\ldots,\mathcal
    D_M$ to condition $M$ \acp{GP}. Furthermore, we condition another \ac{GP} from the data $D_t$ of the target process (cf. Sec.~\ref{sec:problem}). Afterwards, an ensemble of all \acp{GP} is built and used as a simulation in which the optimization takes place. The fundamental idea of this method was first introduced by the authors of~\cite{feurer2018scalable}. In the following, we adapt the method to better suit the application of optimizing manufacturing processes and describe the method in full detail. 
    
    In a first step, we normalize the data of the $M$ related source tasks. For each dataset $\mathcal{D}_m$, we identify the value $\bm x$ that corresponds to the minimum value $y$. Let this value be denoted as $\hat{\bm x}_m$, where
    \begin{equation} \label{eq:min_norm}
   \hat{\bm x}_m = \bm x_{i^\ast}^m \quad \text{ s.t. }\quad i^\ast = \argmin_{i= 1, 2, \ldots, N_m} y_i^m~.
    \end{equation}
    Additionally, we calculate a scaling factor by means of
    \begin{equation}
        \Delta \bm x_m = \left|\max_i\left({\bm x_i^m}\right) - \min_i\left({\bm x_i^m} \right)\right|
    \end{equation}
    in order to obtain $M$ scaled datasets according to
    \begin{equation} \label{eq:03_normalization}
        \tilde{\bm x_i}^m = \frac{\bm x_i^m - \hat{\bm x}_m}{\Delta \bm x_m} ~.
    \end{equation}
    The respective $y$-values are standardized according to
    \begin{equation} \label{eq:03_standardization}
        \tilde{y}_i^m = \frac{y_i^m - \mu_m}{\sigma_m}
    \end{equation}
    with the mean value $\mu_m = \mathrm{Mean}_i(y_i^m)$ and the standard deviation $\sigma_m = \mathrm{Std}_i(y_i^m)$ for each dataset. 
    This leads to normalized datasets $\tilde{\mathcal{D}}_1, \dots, \tilde{\mathcal{D}}_M$ with $\tilde{\mathcal{D}}_m = \{(\tilde{\bm x}_i^m, \tilde{y}_i^m) | i=1,2,\ldots,N_m\}$. 
    These data sets are utilized to condition $M$ \acp{GP} in order to approximate the underlying function to obtain a posterior $f_m(\x | \tilde{D}_m)$, $m=1,\ldots,M$. 
    
    After this initialization, the online optimization starts. Therefore, an initial search point $\bm x^t_0$ is selected to the best of the user's knowledge, and the respective quality $y^t_0$ is measured. This can, for example, be a random guess or a value that has proven to be successful in an earlier attempt. In order to train another \ac{GP} for the target process, at least two data points are required. The most straightforward strategy is to select another random point in the parameter space. In many cases, it is more useful to choose an optimum of a source task as starting point and a second point in the direct vicinity. It has been shown to be beneficial to include valuable domain knowledge in the optimization process~\cite{souza2021optprior}. The resulting $y$-values are measured in order to create the first two instances of the target dataset $\mathcal{D}_t$, i.e. $\mathcal D_t = \{(\x_0^t, y_0^t), (\x_1^t, y_1^t)\}$.
    
    Subsequently, the target dataset $\mathcal{D}_t$ is centralized by means of
    \begin{equation} \label{eq:03_target_normalization}
        \tilde{\bm x}^t_i = \frac{\bm x^t_i - \bm x^t_0}{\Delta \bm x_t}
    \end{equation}
    with $\Delta \x_t = \bm x_\text{max} - \bm x_\text{min}$ to obtain $\tilde{\mathcal{D}}_t$. In contradiction to Eq.~\eqref{eq:03_normalization}, the target data set is normalized around the first value, because we lack the minimum value of Eq.~\eqref{eq:min_norm} in a data set that is continuously built up. Afterwards, the target-\ac{GP} is conditioned by means of $f_t(\bm x|\tilde{D}_t)$. Subsequently, the optimization procedure starts by generating an ensemble from the $M+1$ \acp{GP}. The following steps to obtain a weighted ensemble by means of
    \begin{equation} \label{eq:04_ensemble_function}
        \overline{f}(\x|\mathcal{\tilde{D}}_{1:M}, \mathcal{\tilde{D}}_t) = w_t \cdot f_t(\bm x | \mathcal{\tilde{D}}_t) + \sum_{m=1}^{M} w_m \cdot f_m(\bm x|\mathcal{\tilde{D}}_m)
    \end{equation}
    were proposed in~\cite{feurer2018scalable}. In Eq.~\eqref{eq:04_ensemble_function}, $\bm w = \begin{bmatrix}w_1 & \ldots & w_M & w_t\end{bmatrix} \in \mathds{R}^{M+1}$ are the weights of the individual \acp{GP}, which are obtained by minimizing the loss function
    \begin{equation} \label{eq:03_weight_loss_fn}
    \mathcal{L} = \sum_{m=1}^{M} \mathcal{L}_m + \mathcal{L}_t    
    \end{equation}
    with the loss values
    \begin{equation} \label{eq:04_loss}
        \mathcal{L}_m = \sum_{j=1}^{N_t} \sum_{k=1}^{N_t} \mathbb{1}\bigl\{\left[f_m\left(\bm x_j^t \right) < f_m\left( \bm x_k^t \right) \right]\oplus \left[y_j^t < y_k^t \right]\bigl\} ~.
    \end{equation}
    Here, $\oplus$ denotes the exclusive-or operator and the $\mathbb{1}(\cdot)$-operator counts the number of non-zero elements. The loss $\mathcal{L}_t$ is calculated equivalently to Eq.~\eqref{eq:04_loss}, but with a leave-one-out-cross validation, as it is described in~\cite{feurer2018scalable}.  This is a cross-validation technique, in which a data sample is removed from the \ac{GP} and its prediction quality at the respective spot is evaluated. The benefit of the loss function in Eq.~\eqref{eq:04_loss} is the possibility of comparing the location of optima in the course of the data and not to focus on a mean squared error. The weights are required to satisfy the conditions
    \begin{align}
        \sum_{i=1}^{M+1} w_i = 1 \text{ and } w_i \ge 0 \; \forall i ~.
    \end{align}
    
    In the following, the ensemble function in Eq.~\eqref{eq:04_ensemble_function} is abbreviated by means of $\overline{f}(\bm x|\mathcal{D})$. In laser cutting, we assume that the ground truth function $f(\cdot)$ of  Eq.~\eqref{eq:03_gt_f} shows some similarities across different laser powers and sheet metal thicknesses, apart from some linear transformations. The normalization procedure of the target data set must be applied before the calculation of the loss function in Eq.~\eqref{eq:04_loss} in order to obtain comparable results between all different processes.

    Afterwards, the ensemble function $\overline{f}(\bm x|\mathcal{D})$ is used in a constrained optimization function as in Eq.~\eqref{eq:bo} to find the optimizer $\bm x^\ast$ according to
    \begin{equation} \label{eq:03_minimization}
        \bm x^\ast = \arg \min_{\bm x}  \overline{f}(\bm x|\mathcal{D}) + \lambda \cdot g(\bm x)
    \end{equation}
    subject to $\bm x \in [\bm a, \bm b]$. The found optimizer $\bm x^\ast$ is applied to the target process to measure the respective $y$-value and both are added to the target dataset $\mathcal{D}_t$. This marks the end of one iteration step.  The complete algorithm is given in Alg.~\ref{alg:1}.

\begin{algorithm}[t]
\caption{Pseudo code of the proposed \ac{TLBO} method for machine parameter adaptation.}
\begin{algorithmic}[1]
\State Given data set $\mathcal{D}_1, \dots, \mathcal{D}_M$ 
\State Normalize all datasets according to Eq.~\eqref{eq:03_normalization} and Eq.~\eqref{eq:03_standardization}
\State Acquire initial two data points $\x_0^t$ and $\x_1^t$ and measure the respective product qualities $y_0^t$, $y_1^t$ to initialize $\mathcal{D}_t$
\For{each data set $\tilde{\mathcal D}_m$ with $m=1,\ldots,M$}
    \State Condition \iac{GP} $f_m(\bm x | \tilde{\mathcal D}_m)$    
\EndFor
\State Set $i = 1$
\While{Quality $y_i^t$ not sufficient}
    \State Normalize target data set according to Eq.~\eqref{eq:03_standardization} and Eq.~\eqref{eq:03_target_normalization}
    \State Condition \iac{GP} for the target process $f_t(\bm x | \tilde{\mathcal D}_t)$   
    \State Calculate the weights $\w$ by minimizing Eq.~\eqref{eq:03_weight_loss_fn}
    \State Calculate the minimizer in Eq.~\eqref{eq:03_minimization} to obtain $\bm x_i^t$
    \State Apply $\bm x_i^t$ to the process to obtain $y_i^t$
    \State \begin{varwidth}[t]{\linewidth}{$\mathcal{D}_t \leftarrow \mathcal{D}_t \cup (\bm x_i^t, y_i^t)$\Comment{add $(\x_i^t, y_i^t)$ to target data \par
    set}}
    \end{varwidth}
    \State $i \leftarrow i + 1$
\EndWhile

\end{algorithmic}
\label{alg:1}
\end{algorithm}

 \section{Evaluation} \label{sec:evaluation}
In the following section, we evaluate our algorithm in two similar scenarios. At first, we build a surrogate model based on thermal cutting data to evaluate our method. Afterwards, we present an experiment conducted on an actual thermal laser cutting machine to show the optimization capabilities in a realistic production scenario.

In thermal laser cutting, a laser is focused on a sheet metal to cut its edge. Manufacturers must deal with different types of sheet metal and variability in the quality of raw materials, which complicates the optimization process. The need to adapt to these fluctuating conditions highlights the importance of robust optimization techniques that can perform well even with limited or varying data. Currently, when the type or thickness of the sheet metal changes during production, the adjustment of production parameters is performed manually. This process is labor-intensive and relies heavily on the operator's domain knowledge and experience. In addition, manual adjustments can cause material waste, as several trials are typically required to identify the optimal settings. While optimal parameter configurations are often well known for many different types and thicknesses of steel, this knowledge is rarely used during a new optimization process. 

In the following, we optimize a laser cutting process by manipulating three machine parameters in order to obtain a high quality cut. The parameters to be optimized are the \emph{feedrate} of the laser, i.e., the velocity of the laser to move on the sheet metal, the \emph{gas pressure}, which influences the medium's density and affects the laser's energy absorption, and the \emph{focal position}, which determines the vertical position of the laser's focus. The quality of a cutting edge in sheet cutting can be characterized by multiple features. In the following, we focus on the burr height of the cutting edge, as this property can be estimated solely from image data. Other possible features are roughness, drag lines, or perpendicularity, which can be incorporated in the optimization using 3D measurement systems~\cite{Stahl2024drag} and \ac{SOTA} computer vision techniques~\cite{Tatzel2021_1000137690, stahl2023comprehensive}. 

\subsection{Data Acquisition}
  The tasks in laser cutting differ in several characteristics, namely metal type, sheet thickness, laser power, distance between nozzle and sheet as well as nozzle type. We used data collected for 30 different tasks. The tasks include thicknesses of $\SI{50}{mm}$ -- \SI{150}{mm}, laser powers between \SI{6}{kW} and \SI{12}{kW}, and eight different types of steel. In each task, the adjustable cutting parameters feedrate, gas pressure, and focal position are varied in a grid, and the corresponding burr is estimated using a vision-based approach introduced in \cite{stahl2023comprehensive}. For each data point, a sample rectangle was cut from the metal sheet. For this sample, the burr height was calculated using the sample mean of the four cut edges. The data sets of the source tasks included between 20 and 120 data points with an average of 73 points per task.

\subsection{Thermal Cutting Simulation}
\label{sec:simu_results}
To perform a statistical analysis and a comparison with other methods, we choose one of the 30 previously trained tasks as the target task and train a  \ac{GP} surrogate model on that. For this purpose, we perform standard \ac{GP} regression using a linear combination of an \ac{SE} kernel and a Matern kernel with $\nu=2.5$ \cite{williams2006gaussian}. It was found by performing a hyper-parameter optimization using the \emph{Optuna} framework~\cite{optuna_2019}. The hyper-parameters of the \ac{SE} kernel were identified by performing a gradient-descent search with the training data. 

The surrogate model has to replicate the underlying data generating process as good as possible. In this experiment, we test our optimization on the surrogate model instead of the real hardware. This is necessary since it is highly expensive and time-consuming to cut a huge amount of sheet metal in production. In addition, it allows us to compare our approach with baselines and other \ac{SOTA} methods for several independent trials. With the procedure of calculating the normalization vector described in Eq.~\eqref{eq:03_weight_loss_fn}, we observed the possibility that the target process sometimes was weighted too small compared to the source tasks, because the loss function in Eq.~\eqref{eq:03_weight_loss_fn} only compares tasks by ranking and not by the actual function values. To counteract this behavior, we enforce a linear growth of the target weight $w_t$ according to $w_{t,\mathrm{
        lin}} = \min \left( \alpha_0 \cdot i + \alpha_1, \beta \right) $  at the $i$-th iteration
    with parameters $\alpha_0, \alpha_1 \ge 0$ and a peak weight $\beta > 0$. The forced target weight $w_{t,\mathrm{for}}$ is then calculated according to $w_{t,\mathrm{for}} = \max \left(w_t, w_{t,\mathrm{lin}} \right)$.

Our simulative setup consists of one target task and a subset of the initial 30 tasks as source tasks. In this subset, the sheet thickness and laser power are similar to the target task. We chose only a specific subset of tasks for three reasons. Firstly, it is shown that source tasks that are too different from the target task do not contribute at all to transfer learning or even hinder its performance \cite{zhang2023_negtrans, wang2019_negtrans}. Secondly, the \ac{SOTA} method \ac{MTGP}~\cite{Bonilla2007MTGP} scales poorly with the number of tasks, limiting the number of tasks that can be used during evaluation. Lastly, it is much more realistic from a practical point of view to have data on only a few tasks, since data acquisition is already very elaborate in this domain. 

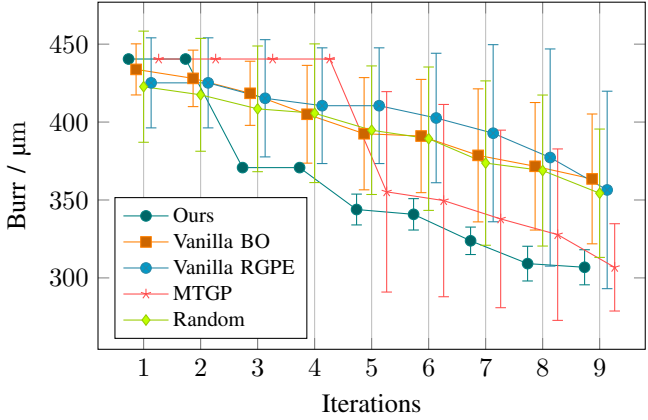
\begin{figure}[tbp]
    \centerline{\begin{tikzpicture}
                \begin{axis}[xtick = {1,2,3,4,5,6,7,8,9},
        xmajorgrids,
        cycle list name=exotic,
        legend pos=south west,
        height=0.7\linewidth,
        width=\linewidth,
        legend cell align={left},
        legend style={nodes={scale=0.8, transform shape}},
        xlabel=Iterations, ylabel=Burr / \si{\micro m}]
            \addplot+[xshift=-0.2cm,legend image post style={xshift=0.2cm}, error bars, y dir=both,y explicit]
            table[x=it, y=cebayopt_mean, col sep=comma, y error=cebayopt_std]{table_results_task_120_ST120MD0-N2H0-30-2_L76_0.4_10000_S355J2.csv};
            \addplot+[xshift=-0.1cm,legend image post style={xshift=0.1cm},error bars, y dir=both,y explicit]
            table[x=it, y=singlegp_mean, col sep=comma, y error=singlegp_std]{table_results_task_120_ST120MD0-N2H0-30-2_L76_0.4_10000_S355J2.csv};
            \addplot+[xshift=0.1cm,legend image post style={xshift=-0.1cm},error bars, y dir=both,y explicit]
            table[x=it, y=rgpeonly_mean, col sep=comma, y error=rgpeonly_std]{table_results_task_120_ST120MD0-N2H0-30-2_L76_0.4_10000_S355J2.csv};
            \addplot+[xshift=0.2cm,legend image post style={xshift=-0.2cm},error bars, y dir=both,y explicit]
            table[x=it, y=multitask_mean, col sep=comma, y error=multitask_std]{table_results_task_120_ST120MD0-N2H0-30-2_L76_0.4_10000_S355J2.csv};
            \addplot+[error bars, y dir=both,y explicit]
            table[x=it, y=random_mean, col sep=comma, y error=random_std]{table_results_task_120_ST120MD0-N2H0-30-2_L76_0.4_10000_S355J2.csv};
            \legend{Ours, Vanilla \ac{BO}, Vanilla \ac{RGPE}, \ac{MTGP}, Random}
        \end{axis}
    \end{tikzpicture}}
    \caption{Minimal burr height predicted after each \ac{BO} iteration as well as the corresponding standard deviation for every method averaged over ten trials. To improve readability, we shift the data slightly on the x-axis for each method. The optimization trials are performed on a surrogate model.}
\label{fig:simu_exp}
\end{figure}

Random sampling and a vanilla \ac{BO} without the use of prior knowledge serve as baselines. We also include \ac{MTGP} in our comparison, as well as the original \ac{RGPE} algorithm proposed in~\cite{feurer2018scalable}. For every source task, a separate \ac{GP} is fitted to estimate the burr height given the cutting parameters. For all \acp{GP}, we apply the same kernels being used for the surrogate model. All transfer learning based methods have access to the same source task \acp{GP}. 

We allow for ten iterations, i.e., laser cuts, and we average the results over ten independent trials. The target task is characterized by a sheet thickness of \SI{12}{mm}, a laser power of \SI{10}{kW} and distance between the nozzle and the sheet of \SI{4}{mm}. 
Fig.~\ref{fig:simu_exp} shows the minimal average target value after each iteration. Our method leads to the quickest improvement, while the other methods converge rather slowly. After five iterations \ac{BO} using \ac{MTGP} yields almost the same improvement in burr height as our method. However, the optimization with \ac{MTGP} is much slower in our experiments, which leads to further complications regarding real-time capability when deployed on the actual laser cutting machine. Furthermore, the standard deviation of \ac{MTGP} is high compared to our method. If more than ten iterations are performed, all methods can converge to a similar optimum. However, we intentionally focused on the first ten steps to highlight the capabilities of our method to achieve a good optimization result with as little data of the target task as possible to reduce time and material. 

In addition, the same experiment was performed multiple times with a different target task, which varies by a different combination of sheet thickness, laser power, and steel type. The optimization results for these tasks are given in Table~\ref{tab1}, with the tasks denoted by the letters A--F. In all cases our method performs best or almost equal to the best result while the other methods are not consistent over all tasks.

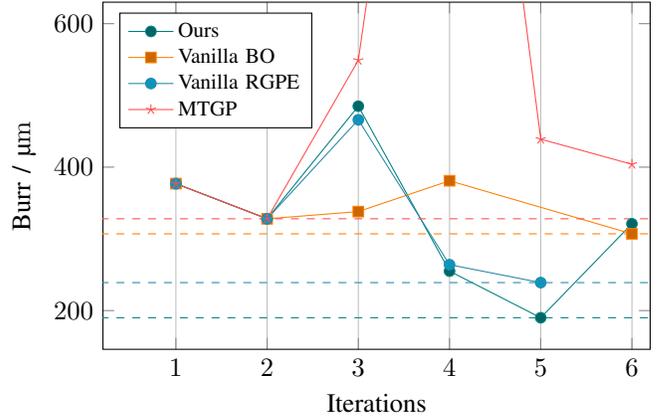
\begin{figure}[tbp]
    \centerline{\begin{tikzpicture}
                \begin{axis}[xtick = {1,2,3,4,5,6},
        xmajorgrids,
        xmax=6.2,
        xmin=0.2,
        ymax=630,
        height=0.7\linewidth,
        width=\linewidth,
        legend cell align={left},
        cycle list name=exotic,
        legend pos=north west,
        legend style={nodes={scale=0.8, transform shape}},
        xlabel=Iterations, ylabel=Burr / \si{\micro m}]
            \addplot+[forget plot, mark=none, domain=0:7, dashed]{190.0};
            \addplot+[]
            table[x=it, y=cebayopt_mean, col sep=comma]{real_hardware_table_results_task_80_ST080MD5-N2S0-30-2_L76_0.4_6000_16Mo3.csv};
            \addplot+[forget plot, mark=none, domain=0:7, dashed]{307.0};
            \addplot+[]
            table[x=it, y=singlegp_mean, col sep=comma]{real_hardware_table_results_task_80_ST080MD5-N2S0-30-2_L76_0.4_6000_16Mo3.csv};
            \addplot+[forget plot, mark=none, domain=0:7, dashed]{239.0};
            \addplot+[]
            table[x=it, y=rgpeonly_mean, col sep=comma]{real_hardware_table_results_task_80_ST080MD5-N2S0-30-2_L76_0.4_6000_16Mo3.csv};
            \addplot+[forget plot, mark=none, domain=0:7, dashed]{328.0};
            \addplot+[]
            table[x=it, y=multitask_mean, col sep=comma]{real_hardware_table_results_task_80_ST080MD5-N2S0-30-2_L76_0.4_6000_16Mo3.csv};
            \legend{Ours, Vanilla \ac{BO}, Vanilla \ac{RGPE}, \ac{MTGP}}    
        \end{axis}
    \end{tikzpicture}}
    \caption{Resulting burr height measured after every optimization step for one optimization run. The optimization was performed on an \SI{8}{mm} steel sheet and a power of \SI{6}{kW}. Dashed lines indicate the minimal burr reached for each method after six iterations. Missing values indicate cut interruptions.}
\label{fig:real_exp}
\end{figure}

\begin{table}[htbp]
\caption{Minimal average burr in \si{\micro m} after ten iterations for ten runs. The processes A--F are characterized by different sheet thicknesses, steel types and laser power. The best method is highlighted in bold and the second best is underlined.}
\begin{center}
\begin{tabular}{l c c c c c}
\toprule

\textbf{Task} & \textbf{Random} & \textbf{Vanilla BO} & \textbf{MTGP} & \textbf{Vanilla RGPE} & \textbf{Ours} \\
\midrule
A & 1028 & 1162 & \underline{890} & 893 & \textbf{849} \\
B & 819 & \textbf{801} & 819 & 809 & \underline{802} \\
C & \underline{354} & 364 & \textbf{307} & 356 & \textbf{307} \\
D & 66 & 84 & \textbf{33} & 76 & \underline{45} \\
E & 330 & 317 & 315 & \underline{276} & \textbf{244} \\
F & 356 & 237 & \underline{232} & 236 & \textbf{185} \\
\bottomrule
\end{tabular}
\label{tab1}
\end{center}
\end{table}

\subsection{Real Hardware Experiment}
On the real hardware, we perform almost the same experiment as described in Sec.~\ref{sec:simu_results} with some minor deviations. For laser cutting, we use a \emph{TruLaser fiber} cutting machine. This time we only perform one optimization run per method due to limited time and metal sheet resources. To save some additional cuts, we use the same two starting points for each method. This also allows for a direct comparison of the optimization procedures alone. As a target process, we use an \SI{8}{mm} steel sheet of alloyed steel 
with a power of \SI{6}{kW}. The results are shown in Fig.~\ref{fig:real_exp}. We see that all methods start with selecting a parameter configuration that performs worse than the two starting points. This is because the starting points already lead to a low burr and in the vicinity of the starting points almost no gain in quality can be achieved. In the fourth and fifth iterations, vanilla RGPE and our method both find parameter configurations with a better burr, while MTGP cannot find an acceptable solution at all. Vanilla BO and vanilla RGPE select configurations that lead to cut interruptions in the fifth and sixth iteration, respectively. In general, our method selects the best configuration that results in a burr height of \SI{190}{\micro m} while avoiding cut interruptions.

    

\section{Conclusion and Future Work}

In this paper, we used an algorithm based on \ac{RGPE} designed for online \ac{BO}, emphasizing minimal reliance on target task data. By incorporating task-specific data scaling, our method ensures better task comparison by transforming data into a comparable regime. Using source task optima as starting points, the algorithm achieves rapid convergence with limited exploration. These features collectively enable highly data-efficient optimization, as demonstrated in our experiments on parameter adaptation for a laser cutting machine. The results highlight the algorithm's capability to minimize burr formation on previously unknown steel sheet types.

Our evaluations, conducted on both a \ac{GP} surrogate model and an actual laser cutting machine, validate the proposed approach. On the surrogate model, our algorithm exhibits fast convergence and performance comparable to or better than \ac{MTGP}, with significantly reduced computational overhead. In real-world experiments, it achieves high-quality cuts with minimal burr in just a few iterations. These findings confirm the practicality and efficiency of our method for industrial parameter optimization tasks.

In the future, we plan to evaluate our method in different industrial applications such as welding tasks and battery manufacturing tasks such as beading and flanging. Further areas of research include the analysis of task similarity, as this is a crucial step in \ac{TLBO}.



\bibliographystyle{IEEEtran}
\bibliography{references.bib}

\begin{thebibliography}{10}
\providecommand{\url}[1]{#1}
\csname url@samestyle\endcsname
\providecommand{\newblock}{\relax}
\providecommand{\bibinfo}[2]{#2}
\providecommand{\BIBentrySTDinterwordspacing}{\spaceskip=0pt\relax}
\providecommand{\BIBentryALTinterwordstretchfactor}{4}
\providecommand{\BIBentryALTinterwordspacing}{\spaceskip=\fontdimen2\font plus
\BIBentryALTinterwordstretchfactor\fontdimen3\font minus \fontdimen4\font\relax}
\providecommand{\BIBforeignlanguage}[2]{{%
\expandafter\ifx\csname l@#1\endcsname\relax
\typeout{** WARNING: IEEEtran.bst: No hyphenation pattern has been}%
\typeout{** loaded for the language `#1'. Using the pattern for}%
\typeout{** the default language instead.}%
\else
\language=\csname l@#1\endcsname
\fi
#2}}
\providecommand{\BIBdecl}{\relax}
\BIBdecl

\bibitem{chia2022process}
H.~Y. Chia, J.~Wu, X.~Wang, and W.~Yan, ``Process parameter optimization of metal additive manufacturing: A review and outlook,'' \emph{Journal of Materials Informatics}, vol.~2, no.~4, pp. N--A, 2022.

\bibitem{LI2020117314}
S.~Li, W.~Gong, L.~Wang, X.~Yan, and C.~Hu, ``Optimal power flow by means of improved adaptive differential evolution,'' \emph{Energy}, vol. 198, p. 117314, 2020.

\bibitem{weichert2019review}
D.~Weichert, P.~Link, A.~Stoll, S.~R{\"u}ping, S.~Ihlenfeldt, and S.~Wrobel, ``A review of machine learning for the optimization of production processes,'' \emph{The International Journal of Advanced Manufacturing Technology}, vol. 104, no.~5, pp. 1889--1902, 2019.

\bibitem{dou2023machine}
B.~Dou, Z.~Zhu, E.~Merkurjev, L.~Ke, L.~Chen, J.~Jiang, Y.~Zhu, J.~Liu, B.~Zhang, and G.-W. Wei, ``Machine learning methods for small data challenges in molecular science,'' \emph{Chemical Reviews}, vol. 123, no.~13, pp. 8736--8780, 2023.

\bibitem{kraljevski2023machinelearningsmalldata}
\BIBentryALTinterwordspacing
I.~Kraljevski, Y.~C. Ju, D.~Ivanov, C.~Tschöpe, and M.~Wolff, ``How to do machine learning with small data? -- a review from an industrial perspective,'' 2023. [Online]. Available: \url{https://arxiv.org/abs/2311.07126}
\BIBentrySTDinterwordspacing

\bibitem{xu2023small}
P.~Xu, X.~Ji, M.~Li, and W.~Lu, ``Small data machine learning in materials science,'' \emph{npj Computational Materials}, vol.~9, no.~1, p.~42, 2023.

\bibitem{zhuang2021tlsurvey}
F.~Zhuang, Z.~Qi, K.~Duan, D.~Xi, Y.~Zhu, H.~Zhu, H.~Xiong, and Q.~He, ``A comprehensive survey on transfer learning,'' \emph{Proceedings of the IEEE}, vol. 109, no.~1, pp. 43--76, 2021.

\bibitem{garnett_bayesoptbook_2023}
R.~Garnett, \emph{{Bayesian Optimization}}.\hskip 1em plus 0.5em minus 0.4em\relax Cambridge University Press, 2023.

\bibitem{sano2020application}
S.~Sano, T.~Kadowaki, K.~Tsuda, and S.~Kimura, ``Application of bayesian optimization for pharmaceutical product development,'' \emph{Journal of Pharmaceutical Innovation}, vol.~15, pp. 333--343, 2020.

\bibitem{bommasani2021opportunities}
R.~Bommasani, D.~A. Hudson, E.~Adeli, R.~Altman, S.~Arora, S.~von Arx, M.~S. Bernstein, J.~Bohg, A.~Bosselut, E.~Brunskill \emph{et~al.}, ``On the opportunities and risks of foundation models,'' \emph{arXiv preprint arXiv:2108.07258}, 2021.

\bibitem{subramanian2024towards}
S.~Subramanian, P.~Harrington, K.~Keutzer, W.~Bhimji, D.~Morozov, M.~W. Mahoney, and A.~Gholami, ``Towards foundation models for scientific machine learning: Characterizing scaling and transfer behavior,'' \emph{Advances in Neural Information Processing Systems}, vol.~36, 2024.

\bibitem{zhang2023large}
H.~Zhang, S.~S. Dereck, Z.~Wang, X.~Lv, K.~Xu, L.~Wu, Y.~Jia, J.~Wu, Z.~Long, W.~Liang \emph{et~al.}, ``Large scale foundation models for intelligent manufacturing applications: a survey,'' \emph{arXiv preprint arXiv:2312.06718}, 2023.

\bibitem{feurer2018scalable}
M.~Feurer, B.~Letham, and E.~Bakshy, ``Scalable meta-learning for bayesian optimization using ranking-weighted gaussian process ensembles,'' in \emph{AutoML Workshop at ICML}, vol.~7, 2018, p.~5.

\bibitem{Bonilla2007MTGP}
E.~V. Bonilla, K.~Chai, and C.~Williams, ``Multi-task gaussian process prediction,'' in \emph{Advances in Neural Information Processing Systems}, J.~Platt, D.~Koller, Y.~Singer, and S.~Roweis, Eds., vol.~20.\hskip 1em plus 0.5em minus 0.4em\relax Curran Associates, Inc., 2007.

\bibitem{swersky2013_MTGPBO}
K.~Swersky, J.~Snoek, and R.~P. Adams, ``Multi-task bayesian optimization,'' in \emph{Advances in Neural Information Processing Systems}, C.~Burges, L.~Bottou, M.~Welling, Z.~Ghahramani, and K.~Weinberger, Eds., vol.~26.\hskip 1em plus 0.5em minus 0.4em\relax Curran Associates, Inc., 2013.

\bibitem{tighineanu2022transfer}
P.~Tighineanu, K.~Skubch, P.~Baireuther, A.~Reiss, F.~Berkenkamp, and J.~Vinogradska, ``Transfer learning with gaussian processes for bayesian optimization,'' in \emph{International Conference on Artificial Intelligence and Statistics}.\hskip 1em plus 0.5em minus 0.4em\relax PMLR, 2022, pp. 6152--6181.

\bibitem{golovin2017google}
D.~Golovin, B.~Solnik, S.~Moitra, G.~Kochanski, J.~Karro, and D.~Sculley, ``Google vizier: A service for black-box optimization,'' in \emph{Proceedings of the 23rd ACM SIGKDD international conference on knowledge discovery and data mining}, 2017, pp. 1487--1495.

\bibitem{Xilo2023survey}
X.~Wang, Y.~Jin, S.~Schmitt, and M.~Olhofer, ``Recent advances in bayesian optimization,'' \emph{ACM Comput. Surv.}, vol.~55, no. 13s, Jul. 2023.

\bibitem{bai2023tlbosurvey}
\BIBentryALTinterwordspacing
T.~Bai, Y.~Li, Y.~Shen, X.~Zhang, W.~Zhang, and B.~Cui, ``Transfer learning for bayesian optimization: A survey,'' 2023. [Online]. Available: \url{https://arxiv.org/abs/2302.05927}
\BIBentrySTDinterwordspacing

\bibitem{Tatzel2021_1000137690}
L.~F. Tatzel, ``\BIBforeignlanguage{german}{Verbesserungen beim laserschneiden mit methoden des maschinellen lernens},'' Ph.D. dissertation, Karlsruher Institut für Technologie (KIT), 2021.

\bibitem{Menold2024laserBO}
T.~Menold, V.~Onuseit, M.~Buser, M.~Haas, N.~Bär, and A.~Michalowski, ``Laser material processing optimization using bayesian optimization: a generic tool,'' \emph{Light: Advanced Manufacturing}, vol.~5, no. LAM2023110081, p. 355, 2024.

\bibitem{michalowski2023advanced}
A.~Michalowski, A.~Ilin, A.~Kroschel, S.~Karg, P.~Stritt, A.~Dais, S.~Becker, G.~Kunz, S.~Sonntag, M.~Lustfeld, P.~Tighineanu, V.~Onuseit, M.~Haas, T.~Graf, and H.~Ridderbusch, ``Advanced laser processing and its optimization with machine learning,'' in \emph{Laser Applications in Microelectronic and Optoelectronic Manufacturing (LAMOM) XXVIII}, A.~Narazaki, L.~Gemini, and J.~Kleinert, Eds., vol. 12408, International Society for Optics and Photonics.\hskip 1em plus 0.5em minus 0.4em\relax SPIE, 2023, p. 1240803.

\bibitem{stahl2023comprehensive}
J.~Stahl, A.~Frommknecht, and M.~Huber, ``Comprehensive quantitative quality assessment of thermal cut sheet edges using convolutional neural networks.'' in \emph{BMVC}, 2023, pp. 480--481.

\bibitem{Stahl2024drag}
J.~Stahl, S.~Zengl, A.~Frommknecht, C.~Jauch, and M.~F. Huber, ``Algorithmic assessment of drag on thermally cut sheet metal edges,'' \emph{tm - Technisches Messen}, 2024.

\bibitem{stahl2019quick}
J.~Stahl and C.~Jauch, ``Quick roughness evaluation of cut edges using a convolutional neural network,'' in \emph{Fourteenth International Conference on Quality Control by Artificial Vision}, vol. 11172.\hskip 1em plus 0.5em minus 0.4em\relax SPIE, 2019, pp. 175--181.

\bibitem{leiner2023cut}
K.~Leiner, F.~P. Dollmann, M.~F. Huber, M.~Geiger, and S.~Leinberger, ``Cut interruption detection in the laser cutting process using rocket on audio signals,'' in \emph{2023 IEEE 21st International Conference on Industrial Informatics (INDIN)}.\hskip 1em plus 0.5em minus 0.4em\relax IEEE, 2023, pp. 1--6.

\bibitem{murphy2012machine}
K.~P. Murphy, \emph{Machine learning: a probabilistic perspective}.\hskip 1em plus 0.5em minus 0.4em\relax MIT press, 2012.

\bibitem{souza2021optprior}
A.~Souza, L.~Nardi, L.~B. Oliveira, K.~Olukotun, M.~Lindauer, and F.~Hutter, ``Bayesian optimization with a prior for the optimum,'' in \emph{Machine Learning and Knowledge Discovery in Databases. Research Track}, N.~Oliver, F.~P{\'e}rez-Cruz, S.~Kramer, J.~Read, and J.~A. Lozano, Eds.\hskip 1em plus 0.5em minus 0.4em\relax Cham: Springer International Publishing, 2021, pp. 265--296.

\bibitem{williams2006gaussian}
C.~K. Williams and C.~E. Rasmussen, \emph{Gaussian processes for machine learning}.\hskip 1em plus 0.5em minus 0.4em\relax MIT press Cambridge, MA, 2006, vol.~2, no.~3.

\bibitem{optuna_2019}
T.~Akiba, S.~Sano, T.~Yanase, T.~Ohta, and M.~Koyama, ``Optuna: A next-generation hyperparameter optimization framework,'' in \emph{Proceedings of the 25th {ACM} {SIGKDD} International Conference on Knowledge Discovery and Data Mining}, 2019.

\bibitem{zhang2023_negtrans}
W.~Zhang, L.~Deng, L.~Zhang, and D.~Wu, ``A survey on negative transfer,'' \emph{IEEE/CAA Journal of Automatica Sinica}, vol.~10, no.~2, pp. 305--329, 2023.

\bibitem{wang2019_negtrans}
Z.~Wang, Z.~Dai, B.~Póczos, and J.~Carbonell, ``Characterizing and avoiding negative transfer,'' in \emph{2019 IEEE/CVF Conference on Computer Vision and Pattern Recognition (CVPR)}, 2019, pp. 11\,285--11\,294.

\end{thebibliography}

\end{document}